\documentclass[10pt,twocolumn,letterpaper]{article}

\usepackage{cvpr}              

\usepackage[dvipsnames]{xcolor}
\usepackage{overpic}
\usepackage{enumitem} 
\usepackage{overpic} 
\usepackage{color}
\usepackage{array}
\usepackage{bm}
\usepackage{multirow}
\usepackage{multicol}
\usepackage{graphicx}
\usepackage{amsmath}
\usepackage{amssymb}
\usepackage{booktabs}
\usepackage{cuted}
\usepackage{capt-of}
\usepackage{algorithm2e}

\newcommand{\ignorethis } [1] {}

\definecolor{turquoise}{cmyk}{0.65,0,0.1,0.3}
\definecolor{purple}{rgb}{0.65,0,0.65}
\definecolor{dark_green}{rgb}{0, 0.5, 0}
\definecolor{orange}{rgb}{0.8, 0.6, 0.2}
\definecolor{red}{rgb}{0.8, 0.2, 0.2}
\definecolor{darkred}{rgb}{0.6, 0.1, 0.05}
\definecolor{blueish}{rgb}{0.0, 0.3, .6}
\definecolor{light_gray}{rgb}{0.7, 0.7, .7}
\definecolor{pink}{rgb}{1, 0, 1}
\definecolor{greyblue}{rgb}{0.25, 0.25, 1}

\newcommand{\rfirst}[1]{ \textbf{#1} }
\newcommand{\rsecond}[1]{ \underline{#1} }

\newcommand{\zhengfei}[1]{}
\newcommand{\fujun}[1]{}
\newcommand{\sai}[1]{}
\newcommand{\ZS}[1]{}

\definecolor{color1}{rgb}{0.36470588235, 0.7294117647, 0.43921568627}
\definecolor{color2}{rgb}{0.88235294117, 0.27450980392, 0.2431372549}
\definecolor{color3}{rgb}{0.58823529411, 0.46666666666, 0.4}
\definecolor{color4}{rgb}{0.30196078431, 0.62352941176, 0.88235294117}
\definecolor{color5}{rgb}{0.8862745098, 0.66274509803, 0.22745098039}

\newcommand{\method}{\textbf{Buffer Anytime}\xspace}
\newcommand{\comment}[1]{}

\definecolor{cvprblue}{rgb}{0.21,0.49,0.74}
\usepackage[pagebackref,breaklinks,colorlinks,citecolor=cvprblue]{hyperref}

\def\paperID{4189} 
\def\confName{CVPR}
\def\confYear{2025}

\title{\method: Zero-Shot Video Depth and Normal from Image Priors}

\author{
Zhengfei Kuang\textsuperscript{1}, Tianyuan Zhang\textsuperscript{2}, Kai Zhang\textsuperscript{3}, Hao Tan\textsuperscript{3}, Sai Bi\textsuperscript{3}, Yiwei Hu\textsuperscript{3}, \\
Zexiang Xu\textsuperscript{3}, Milos Hasan\textsuperscript{3}, Gordon Wetzstein\textsuperscript{1}, Fujun Luan\textsuperscript{3}\\
\textsuperscript{1}Stanford University\qquad 
\textsuperscript{2}Massachusetts Institute of Technology\qquad 
\textsuperscript{3}Adobe Research\\
{\tt\small \{zhengfei,gordonwz\}@stanford.edu \qquad tianyuan@mit.edu}\\
{\tt\small \{kaiz,hatan,sbi,yiwhu,zexu,mihasan,fluan\}@adobe.com}\\
{\tt\small \url{bufferanytime.github.io}}\\
}
\begin{document}
\maketitle

\begin{strip}\centering
\vspace*{-50pt}
\includegraphics[width=1\textwidth]{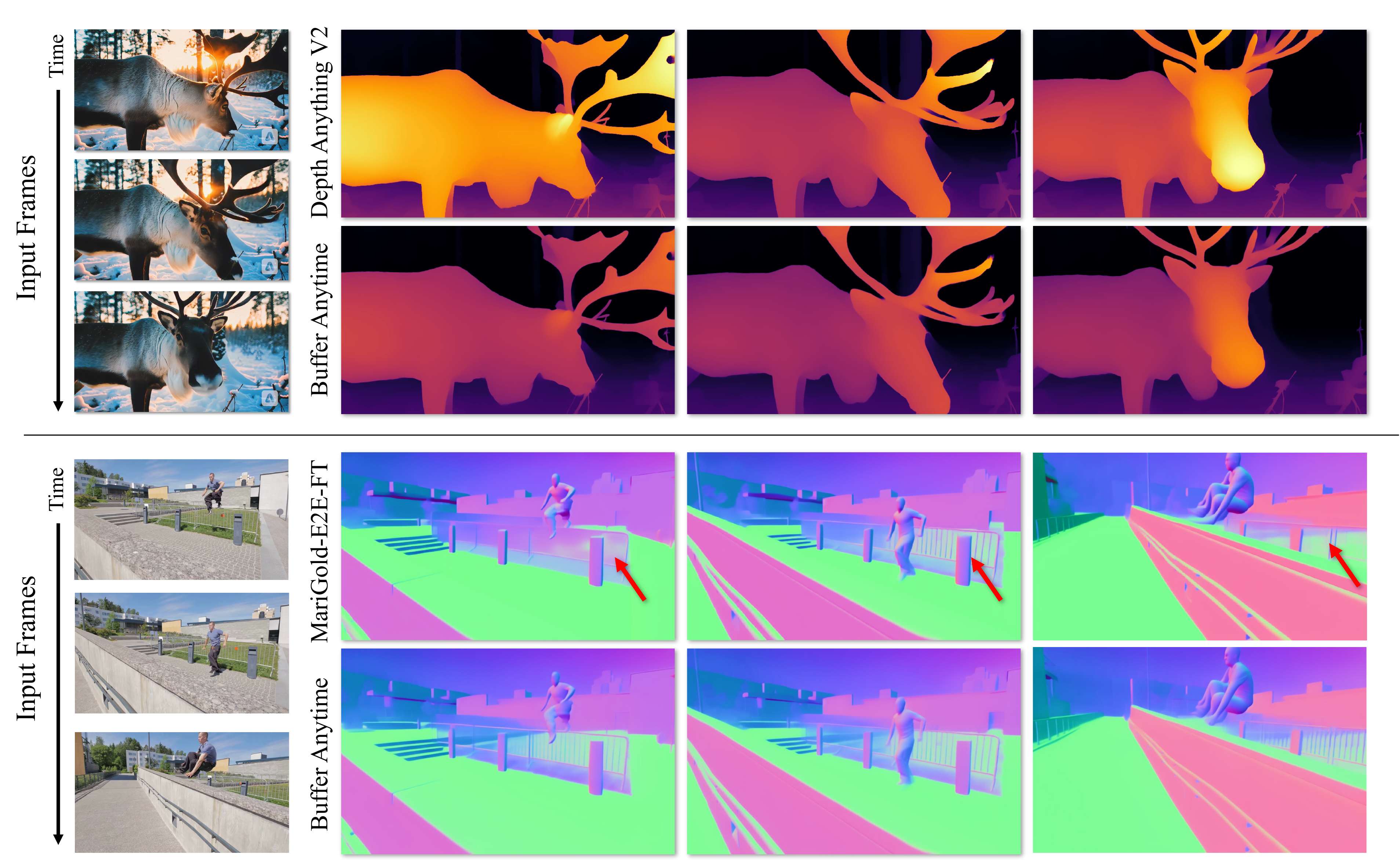}
\vspace*{-10pt}
\captionof{figure}{\method improves temporal consistency in video geometry estimation without paired training data. Top: Comparison of depth estimation between Depth Anything V2~\cite{DepthAnythingV2} and our method on a challenging dynamic scene with lighting variations. While the original model shows inconsistent depth predictions across frames, our approach maintains stable depth estimates. Bottom: Surface normal estimation comparison between Marigold-E2E-FT~\cite{E2EFT} and our method on an outdoor scene with complex geometry. Our method preserves consistent normal maps across frames while maintaining accurate geometric details. In both cases, our method achieves better temporal consistency without requiring video--geometry paired training data. }
\vspace*{-3pt}
\end{strip} 
\begin{abstract}
We present \method, a framework for estimation of depth and normal maps (which we call geometric buffers) from video that eliminates the need for paired video--depth and video--normal training data. Instead of relying on large-scale annotated video datasets, we demonstrate high-quality video buffer estimation by leveraging single-image priors with temporal consistency constraints. Our zero-shot training strategy combines state-of-the-art image estimation models based on optical flow smoothness through a hybrid loss function, implemented via a lightweight temporal attention architecture. Applied to leading image models like Depth Anything V2 and Marigold-E2E-FT, our approach significantly improves temporal consistency while maintaining accuracy. Experiments show that our method not only outperforms image-based approaches but also achieves results comparable to state-of-the-art video models trained on large-scale paired video datasets, despite using no such paired video data.
\end{abstract}    
\section{Introduction}
\label{sec:intro}

Acquiring depth and normal maps from monocular RGB input frames has been a fundamental research topic in computer vision for decades. Serving as a bridge between 2D images and 3D representations, advances in this field have enabled breakthrough applications across domains such as embodied AI, 3D/4D reconstruction and generation, and autonomous driving. 

Recent advances in foundational models, including image/video diffusion models~\cite{StableDiffusionV2, AnimateDiff, SVD, brooks2024sora, po2024state} and large language models (LLMs)~\cite{floridi2020gpt, touvron2023llama}, has accelerated the development of powerful models for image and video buffer estimation. 
By \emph{buffers} we mean information such as per-pixel depth, normals, lighting, or material properties; in this paper we specifically focus on depth and normals (i.e., geometry buffers). 
Empowered by large-scale datasets captured in synthetic environments and the real world, recent works~\cite{MariGold, fu2025geowizard, zeng2024rgb} have demonstrated impressive results in predicting various types of buffers from images.
A promising line of recent work~\cite{ChronoDepth, DepthCrafter} further extends the use of large-scale models for video buffer prediction, showing superior video depth predictions with high fidelity and consistency across frames.

Our work originates from the following question: \textit{Can image-based buffer estimation models help with the task of video buffer estimation?} Comparing to mainstream image/video generative models that take input in lower dimensions as conditions (i.e., text or a single frame) and generate higher-dimensional outputs (i.e., image or video), the buffer estimation models are usually conditioned on RGB image/video of the same size as the desired results; the input already contains rich structural and semantic information. As a result, image inversion models are much more likely to produce consistent contents given similar input conditions compared to text-to-image/video generation models. This observation drives us to explore the possibility of upgrading existing image models for video buffer generation.

In this paper, we demonstrate a positive answer for the above question by showing an effective video geometry buffer model trained from image priors without any supervision from ground truth video geometry data. We propose \method, a flexible zero-shot\footnote{In this context, ``zero-shot'' refers to training without paired video--geometry ground truth data, rather than the traditional meaning of handling unseen classes.} training strategy that combines the knowledge of an image geometry model with existing optical flow methods to ensure both temporal consistency and accuracy of the learned model predictions. We apply the training scheme on two state-of-the-art image models, Depth Anything V2~\cite{DepthAnythingV2} (for depth estimation) and Marigold-E2E-FT~\cite{E2EFT} (for normal estimation), and show significant improvements in different video geometry estimation evaluations. We summarize the major contributions of our work as:

\begin{itemize}
\item A zero-shot training scheme to fine-tune an image geometric buffer model for video geometric buffer generation;
\item A hybrid training supervision that consists of a regularization loss from the image model and an optical flow based smoothness loss;
\item A lightweight temporal attention based architecture for video temporal consistency;
\item Our proposed models outperform the image baseline models by a large margin and are comparable to state-of-the-art video models trained on paired video data.
\end{itemize}

\section{Related Work}
\label{sec:related_works}

Our work intersects with several active research areas in computer vision. We first review recent advances in monocular depth and normal estimation, particularly focusing on large-scale and diffusion-based approaches, and video methods with their efforts in maintaining temporal consistency. We then examine video diffusion models that inspire our temporal modeling strategy.

\subsection{Monocular Depth Estimation}

Monocular depth estimation has evolved through several paradigm shifts. Early works~\cite{hoiem2007recovering, liu2008sift, saxena2008make3d} relied on hand-crafted features and algorithms, while subsequent deep-learning methods \cite{MonoDepth, DoRN, yin2019enforcing, zhang2022hierarchical} improved performance through learned representations. MiDaS~\cite{MiDaS} and Depth Anything~\cite{DepthAnything} further advanced the field by leveraging large-scale datasets, with Depth Anything V2~\cite{DepthAnythingV2} enhancing robustness through pseudo depth labels. Recent works incorporating diffusion models~\cite{MariGold, fu2025geowizard, he2024lotus} achieved fine-grained detail, while E2EFT~\cite{E2EFT} improved efficiency through single-step inference and end-to-end fine-tuning.  While these advances represented major progress for single-image depth estimation, they did not address temporal consistency in videos. NVDS~\cite{NVDS} tackles this challenge by introducing a stabilization framework and video depth dataset, followed by ChronoDepth~\cite{ChronoDepth} and DepthCrafter~\cite{DepthCrafter} which leveraged video diffusion models (e.g. Stable Video Diffusion~\cite{SVD}) to boost prediction quality. Unlike these approaches that require annotated datasets, our method achieves comparable results without ground truth depth maps.

\subsection{Monocular Surface Normal Estimation}
Surface normal estimation has evolved significantly since the pioneering work of Hoiem et al. \cite{hoiem2005automatic, hoiem2007recovering}, who introduced learning-based approaches using handcrafted features. The advent of deep learning sparked numerous neural network-based approaches~\cite{fouhey2014unfolding, eigen2015predicting, wang2015designing, bansal2016marr, wang2020vplnet, do2020surface, bae2021estimating}. Recent advances include Omnidata\cite{eftekhar2021omnidata} and its successor Omnidata v2~\cite{kar20223d}, which leverage large-scale diverse datasets with sophisticated 3D data augmentation. Another line of research~\cite{qi2018geonet, fu2025geowizard, E2EFT, he2024lotus} focuses on jointly predicting normal and depth maps in a unified framework to enforce cross-domain consistency. Recently, DSINE~\cite{DSINE} enhanced robustness by incorporating geometric inductive bias into data-driven methods. However, these works focus solely on single-image predictions without addressing temporal coherence.

\subsection{Video Diffusion Models}
Recent advances in video diffusion models enable high-quality video generation from multimodal input conditions, such as text~\cite{AnimateDiff, brooks2024sora, singer2022make, polyak2024movie}, image~\cite{SVD, guo2023sparsectrl, xu2024camco}, camera trajectory~\cite{he2024cameractrl, kuang2024cvd} and human pose~\cite{hu2023animateanyone, chang2024magicpose, shao2024human4dit}. Among them, AnimateDiff~\cite{AnimateDiff} introduced plug-and-play motion modules for adding dynamics to the existing image model Stable Diffusion~\cite{StableDiffusionV2}, supporting generalized video generation for various personalized domains. 
These advances inspired our temporal modeling approach, though we differ by leveraging image-based priors and optical-flow models without requiring direct video-level supervision.

\section{Method}
\label{sec:method}

We first formulate our problem as follows: given an input RGB video consisting of $K$ frames, $\bm{I}_{1,...,K} \in \mathbb{R}^{K\times H\times W\times 3}$, we aim to predict the corresponding depth maps $\bm{\mathcal{D}}_{1,...,K}\in \mathbb{R}^{K\times H\times W}$, and surface normal maps $\bm{\mathcal{N}}_{1,...,K}\in \mathbb{R}^{K\times H\times W\times 3}$ represented in camera coordinates. For convenience, we will mainly focus on describing the task of depth estimation without loss of generality. While existing state-of-the-art methods for video depth prediction models are trained from paired datasets, i.e., a set of data pairs $(\bm{I}_{1,...,K}, \bm{\hat{\mathcal{D}}}_{1,...,K})_{1,...,N_{\text{data}}}$ of input frames and ground truth depth maps, our model does not require any paired video datasets, instead only relying on the RGB video data $(\bm{I}_{1,...,K})_{1,...,N_{\text{data}}}$. 

The key insight of our approach is to combine image-based diffusion priors and optical-flow based temporal stabilization control. Given an image depth prediction model $f^{\text{image}}_\theta(\bm{I})=\bm{\mathcal{D}}^{\text{image}}$ trained by large-scale image paired datasets to reconstruct the underlying data prior $p^{\text{image}}( \bm{\mathcal{D}} | \bm{I})$, our goal is to develop an upgraded video model $f^{\text{video}}_\theta$ that is backed by $f^{\text{image}}_\theta$ and able to predict depth maps from videos, i.e., $f^{\text{video}}_\theta(\bm{I}_{1,...,K})=\bm{\mathcal{D}}^{\text{video}}_{1,...,K}$. The prediction of $f^{\text{video}}_\theta$ should satisfy two conditions: First, each frame of the depth prediction $\bm{\mathcal{D}}^{\text{video}}_i$ should accommodate the image data prior $p^{\text{image}}( \bm{\mathcal{D}} | \bm{I})$ and second, the frames of the prediction should be temporally stable and consistent with each other.

\subsection{Training Pipeline}

\begin{figure}[t]
  \centering
  \includegraphics[width=\linewidth]{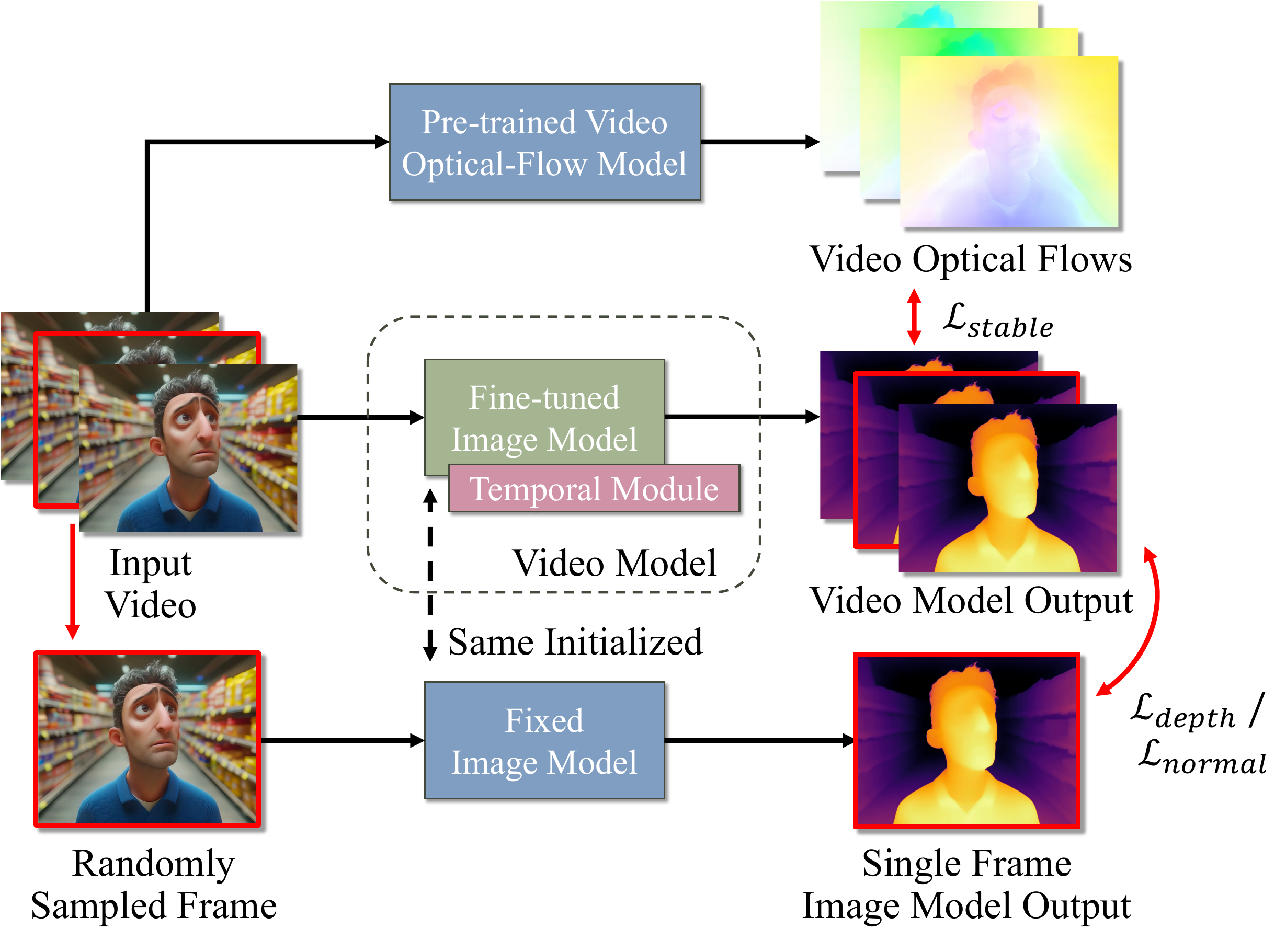}
  \caption{\textbf{Visualization of Our Training Pipeline.} Our pipeline consists of three branches: an optical flow network that extracts optical flow from input video to guide temporal smoothness; a fixed single-frame image model for regularization, and the trained video model that integrates a fine-tuned image backbone with temporal layers. }
  \label{fig:pipeline}
\end{figure}

To achieve the two conditions, we design a novel training strategy (Fig.~\ref{fig:architecture}) that employs two different types of losses: A regularization loss that forces the model to produce results aligned with the image model, and an optical flow based stabilization loss as described in Section~\ref{sec:of}. In depth estimation, the regularization loss is based on the affine-invariant relative loss in previous works~\cite{DepthAnything, E2EFT}: 
\begin{equation}
\mathcal{L}_{\text{depth}} = \frac{1}{HW}||\bm{\hat\mathcal{D}}'_k - \bm{\mathcal{D}}'_k||_2,
\end{equation}
where $H, W$ are the image sizes, $\bm{\mathcal{D}}'_k$ is the predicted depth map of the $k$-th frame normalized by the offset $t=\text{median}(\bm{\mathcal{D}}_k)$ and the scale $s=\frac{1}{HW}\sum_{x} |\bm{\mathcal{D}}_i(x) - t|$, and $\bm{\hat\mathcal{D}}'_k$ is the normalized depth map from the image model. 
In normal estimation, we leverage the latent representation of the backbone model, and simply apply an $L_2$ loss on the predicted latent maps $\bm{z}$:
\begin{equation}
\mathcal{L}_{\text{normal}} = \frac{1}{HW}||\bm{\hat z }_k - \bm{z}_k||_2.
\end{equation}
To speed up training, we randomly select one frame from the video in each iteration and calculate the regularization loss on this frame only. The overall training loss is:
\begin{equation}
    \mathcal{L} = \omega_{\text{reg.}} \cdot \mathcal{L}_{\text{depth / normal}} + \mathcal{L}_{\text{stable}},
\end{equation}
where $\omega_{\text{reg.}}$ is the weight for per-frame regularization with pretrained single-view depth or normal predictors, and is set to $1$ in all experiments, $\mathcal{L}_{\text{stable}}$ is the optical flow based temporal stabilization loss defined in Sec.~\ref{sec:of}. During training, a fixed pre-trained image model and an optical flow model are also deployed aside from the trained video model. We calculate the single frame prediction and the optical flow maps in a just-in-time manner.
For the normal model, calculating the temporal stabilization loss requires decoding the output latent maps into RGB frames first, which is impractical to apply to all frames at once due to memory limitations. Hence we apply the deferred back-propagation technique introduced in~\citet{ARF}. Specifically, we first split the latent map into chunks of 4 frames, then calculate the stabilization loss for each chunk at a time and back-propagate the gradients. We concatenate the gradients of all chunks together as the gradient of the whole latent maps.

\subsection{Optical Flow Based Stabilization}
\label{sec:of}

\begin{figure}[t]
  \centering
  \includegraphics[width=\linewidth]{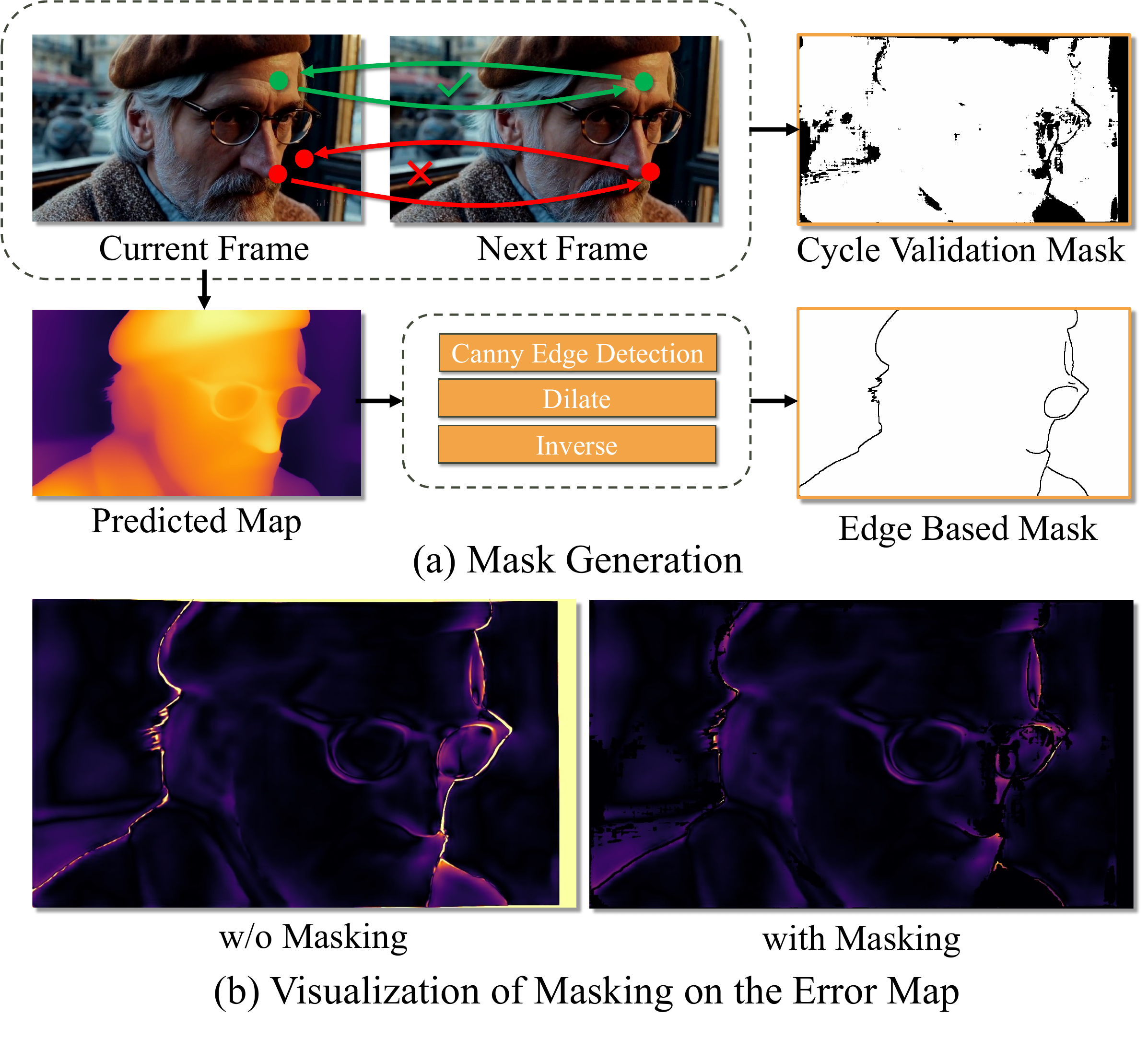}
  \caption{\textbf{Illustration of our masking procedure for the optical flow loss.} \textit{Row 1}: Given two adjacent frames, we first apply cycle validation on the predicted optical flows to filter out the outliers; \textit{Row 2}: We then apply an edge detection procedure on the predicted depth map to mask out the boundaries. \textit{Row 3}: The combination of two masks diminish the effect of inaccurate optical flow prediction to the smoothness error map. }
  \label{fig:of}
\end{figure} 

Single-view image predictors usually suffer from inconsistent results across frames due to the ambiguity of affine transformation (i.e., scale and offset) of the prediction and uncertainty from the model. To alleviate this problem, a reasonable approach is to align the depth predictions between the corresponding pixels across different frames.  
Inspired by previous network prediction stabilizing works~\cite{NVDS, cao2021, wang2022less}, we apply a pre-trained optical flow estimator to calculate the correspondence between adjacent frames for the temporal consistency stabilization. Specifically, given the predicted optical flow maps between two adjacent frames $\bm{I}_k, \bm{I}_{k+1}$ are $\bm{\mathcal{O}}_{k\rightarrow k+1}$ and $\bm{\mathcal{O}}_{k+1\rightarrow k}$, a stabilization loss between the two frames can be defined as:

\begin{align}
\mathcal{L}_{\text{stable}} &= \frac{1}{2HW}\sum_{\bm{x}}|\bm{I}_k(\bm{x}) - \bm{I}_{k+1}(\bm{\mathcal{O}}_{k\rightarrow k+1}(\bm{x}))|_1 \\
&+ \frac{1}{2HW}\sum_{\bm{x}}|\bm{I}_{k+1}(\bm{x}) - \bm{I}_{k}(\bm{\mathcal{O}}_{k+1\rightarrow k}(\bm{x}))|_1 .
\end{align}
In practice, however, the optical flow prediction can be inaccurate or wrong due to the limitations of the pretrained model, harming the effectiveness of the loss as Fig.~\ref{fig:of} shows. To prevent that, we add two filtering methods to curate the correctly corresponded pixels across the frames.
The first method applies the cycle-validation technique that is commonly used in many previous image correspondence methods. Here we only select the pixels in $\bm{I}_k$ that satisfy:
\begin{align}
||\bm{\mathcal{O}}_{k\rightarrow k+1}(\bm{\mathcal{O}}_{k+1\rightarrow k}(\bm{x})) - \bm{x}||_2 \leq \tau_{c},
\end{align}
where $\tau_{c}$ is a hyper-parameter threshold. The second technique is based on the observation that the stabilization loss can be incorrectly overestimated near the boundary areas in the depth frames due to the inaccuracy of the optical flow. Here, our solution is to apply the Canny edge detector~\cite{CANNY} on predicted depth maps, and then filter out the losses on the pixels that are close to the detected edges (i.e., the Manhattan distance is smaller than 3 pixels). The combination of these two filters effectively removes outliers and improves the robustness of our model. 

\subsection{Model Architecture}

\begin{figure*}[h]
  \centering
  \includegraphics[width=\linewidth]{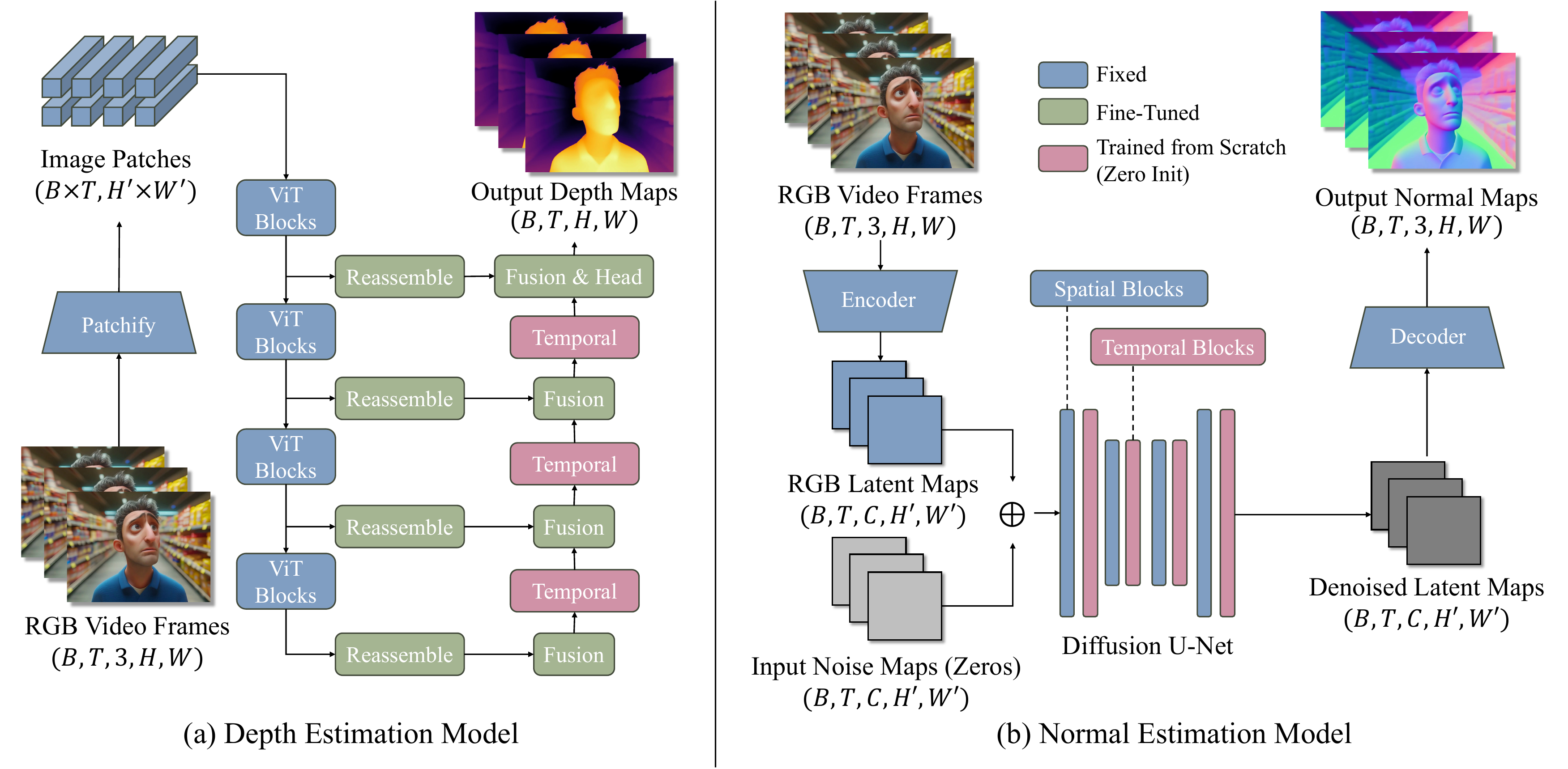}
  \vspace*{-1pt}
  \caption{\textbf{Our Network Architecture.} We present two model architectures for video geometry estimation: (a) A depth estimation model based on Depth Anything V2~\cite{DepthAnythingV2}, where we inject temporal blocks between fusion layers while keeping the ViT backbone frozen. The model processes video frames $(B,T,3,H,W)$ through a patchify layer, multiple ViT blocks with reassemble and fusion operations, and temporal blocks to produce depth maps $(B,T,H,W)$. (b) A normal estimation model built upon Marigold-E2E-FT~\cite{E2EFT}, where we insert temporal blocks between spatial layers in the diffusion U-Net. The model takes RGB video frames as input, processes them through an encoder to obtain latent maps, combines them with zero noise maps, and processes through the U-Net with alternating spatial and temporal blocks to generate normal maps $(B,T,3,H,W)$. Blue blocks are fixed during training, green blocks are fine-tuned, and pink blocks are trained from scratch with zero initialization.}
  \label{fig:architecture}
\end{figure*}

For generating consistent and high-fidelity video results across frames, choosing a powerful and stable image-based backbone model that robustly preserves the structure of the input frames is crucial: it can greatly reduce the inconsistency and ambiguity of the image results, facilitating our video model training process. Recent advances in large-scale image-to-depth and image-to-normal models have brought up many powerful candidates. In this work, we opt to use Depth Anything V2~\cite{DepthAnythingV2} for the backbone of our depth prediction model, and Marigold-E2E-FT~\cite{E2EFT} for our normal prediction model. 

Depth Anything V2~\cite{DepthAnythingV2} is a Dense Vision Transformer (DPT)~\cite{DPT} that consists of a Vision Transformer (ViT)~\cite{ViT} as an encoder, and a lightweight refinement network that fuses the feature outputs of several ViT blocks together and produces the final results. Here we freeze the ViT backbone and only fine-tune the refinement network. As Fig.~\ref{fig:architecture} (a) shows, we inject three temporal blocks in between the fusion blocks, as a bridge to connect the latent maps across different frames. Notice that the ViT blocks are fully detached from the gradient flow, which helps reducing the memory cost during training, enabling support for longer videos.

Marigold-E2E-FT~\cite{E2EFT} is a Latent Diffusion Model~\cite{LDM} built upon Stable-Diffusion V2.0~\cite{StableDiffusionV2}. As Fig.~\ref{fig:architecture} (b) demonstrates, we insert temporal layers between the spatial layers. The original U-Net layers and the autoencoder are both fixed during training. The temporal blocks in both models are structured similarly to the blocks in AnimateDiff~\cite{AnimateDiff}, consisting of several temporal attention blocks followed by a projection layer. The final projection layer of each block is zero-initialized to ensure the model acts the same as the image model when training begins.

Our framework combines the power of state-of-the-art single-view models with temporal consistency through a carefully designed training strategy and architecture. The optical flow based temporal stabilization and regularization losses work together to ensure both high-quality per-frame predictions and temporal coherence, while our lightweight temporal blocks enable efficient processing of video sequences. By freezing the backbone networks and only training the temporal components, we maintain the strong geometric understanding capabilities of the pretrained models while adding temporal reasoning abilities. In the following section, we conduct extensive experiments to validate our design choices and demonstrate the effectiveness of our approach across various video geometry estimation tasks.
\section{Experiments}
\begin{table*}[t]
\centering

{\footnotesize \begin{tabular}{lc ccc c ccc c ccc}
\toprule
\multirow{2}{*}{Method} & \multirow{2}{*}{Time } & \multicolumn{3}{c}{ScanNet~\cite{Scannet}}  && \multicolumn{3}{c}{KITTI~\cite{Kitti}}  &&  \multicolumn{3}{c}{Bonn~\cite{Bonn}} \\ 
\cline{3-5} \cline{7-9} \cline{11-13} 
& & AbsRel $\downarrow$  & $\delta_1$ $\uparrow$ & OPW $\downarrow$ && AbsRel $\downarrow$ & $\delta_1$$\uparrow$ & OPW $\downarrow$ && AbsRel $\downarrow$  & $\delta_1$ $\uparrow$ & OPW $\downarrow$  \\
\midrule

ChronoDepth~\citep{ChronoDepth} & 106s &
0.159  & 0.783 & 0.092 &&
0.151$^*$  & 0.797$^*$ & 0.050 && 
0.109$^*$  & 0.886$^*$ & 0.035 \\

NVDS~\citep{NVDS} & 283s &
0.187  & 0.677 & 0.143 &&
0.253  & 0.588 & 0.089 && 
0.210$^*$  & 0.693$^*$ & 0.068 \\

DepthCrafter~\citep{DepthCrafter} & 270s &
\rsecond{0.125}  & \rsecond{0.848} & \rsecond{0.082} &&
\rfirst{0.110}  & \rfirst{0.881} & 0.111 && 
\rfirst{0.075}  & \rfirst{0.971} & \rsecond{0.029} \\

\midrule

MariGold~\citep{MariGold} & 475s &
0.166  & 0.769 & 0.241 &&
0.149  & 0.796 & 0.235 && 
0.091  & \rsecond{0.931} & 0.109  \\




MariGold-E2E-FT~\citep{E2EFT} & 72s &
0.150  & 0.802 & 0.145 &&
0.151  & 0.779 & 0.100 && 
\rsecond{0.090}  & 0.921 & 0.053 \\

Depth Anything V2~\citep{DepthAnythingV2} & 31s &
0.135  & 0.822 & 0.121 &&
0.140  & 0.804 & \rsecond{0.089} && 
0.119$^*$  & 0.875$^*$ & 0.059 \\

\midrule
Ours (Depth Anything V2) & 33s &
\rfirst{0.123}  & \rfirst{0.853} & \rfirst{0.076} &&
\rsecond{0.119}  & \rsecond{0.865} & \rfirst{0.038} && 
0.102  & 0.925 & \rfirst{0.028} \\

\bottomrule
\end{tabular}}


\caption{\textbf{Evaluation on video depth estimation.} We compare our method against both video-based methods (top section) trained with video supervision and single-image methods (middle section) across three datasets: ScanNet (indoor static), KITTI (outdoor), and Bonn (indoor dynamic). We report absolute relative error (AbsRel), accuracy within 25\% of ground truth ($\delta_1$), and temporal consistency (OPW). Inference time is normalized by frame resolution for fair comparison. *Values marked with asterisk show slight differences from those reported in DepthCrafter. Best results are in \textbf{bold}, second best are \underline{underlined}.}

\label{tab:depth}

\end{table*}

\begin{table*}[t]
\centering

{\small \begin{tabular}{l cccc c cccc }
\toprule
\multirow{2}{*}{Method} & \multicolumn{4}{c}{Sintel~\cite{Sintel}}  && \multicolumn{4}{c}{ScanNet~\cite{Scannet}} \\
\cline{2-5} \cline{7-10}
& Mean $\downarrow$ & Median $\downarrow$ & $11.25^\circ$ $\uparrow$ & OPW $\downarrow$ && Mean $\downarrow$ & Median $\downarrow$  &  $11.25^\circ$ $\uparrow$ & OPW $\downarrow$   \\
\midrule

DSINE~\citep{DSINE} &
35.710  & 29.707 & 20.005 & \rsecond{0.131} &&
22.754 & 11.438 & 55.095 & 0.119 \\

Lotus~\citep{he2024lotus}  
& 34.462 & 28.576 & 19.966 & 0.169 && 
 22.110 & 10.622 & 57.931 & 0.122 \\
 
Marigold~\citep{MariGold} 
& 38.142 & 32.696 & 16.101 & 0.427 && 
23.530 & 11.456 & 55.905 & 0.190  \\


Marigold-E2E-FT ~\citep{E2EFT} &
\rfirst{32.858} & \rfirst{27.300} & \rfirst{22.476} & 0.152 &&
\rsecond{21.587} & \rsecond{9.951} & \rsecond{59.500} & \rsecond{0.092} \\
\midrule
Ours (Marigold-E2E-FT) &
\rsecond{33.421} & \rsecond{28.320} & \rsecond{20.664} & \rfirst{0.065} &&
\rfirst{21.500} & \rfirst{9.915} & \rfirst{59.634} & \rfirst{0.069} \\
\bottomrule
\end{tabular}}

\caption{\textbf{Evaluation on the video normal estimation.} We evaluate on Sintel (synthetic dynamic scenes) and ScanNet (real indoor scenes) datasets. The Mean and Median metrics measure angular error in degrees, $11.25^\circ$ shows percentage of predictions within $11.25^\circ$ of ground truth, and OPW measures temporal consistency. Our method maintains comparable per-frame accuracy while significantly improving temporal stability (OPW) compared to previous approaches. Best results are in \textbf{bold}, second best are \underline{underlined}.}

\label{tab:normal}

\end{table*}

\begin{table}[t]
\centering

{\small \begin{tabular}{l ccc}
\toprule
Method & AbsRel $\downarrow$  & $\delta_1$ $\uparrow$ & OPW $\downarrow$  \\
\midrule
Ours $w_{\text{reg.}}=0.1$ & 
0.137 & 0.846 & 0.040 \\
Ours $w_{\text{reg.}}=3$ & 
0.122 & 0.860 & 0.040 \\
Ours no mask & 
0.122 & 0.860 & \rsecond{0.036} \\
Ours all frames & 
 0.120 & \rfirst{0.865} & \rfirst{0.035} \\
Ours & 
 \rfirst{0.119} & \rfirst{0.865} & 0.038 \\
\bottomrule
\end{tabular}}

\caption{\textbf{Ablation Study on KITTI depth estimation.} We evaluate different variants of our model: different regularization weights ($w_{\text{reg.}}$), removing optical flow correspondence masking (no mask), and using all frames instead of a single frame for regularization (all frames). Our full model achieves the best AbsRel while maintaining strong performance in $\delta_1$ and temporal consistency (OPW).}
\label{tab:ablation}

\end{table}

\begin{figure*}[t]
  \centering
  \includegraphics[width=\linewidth]{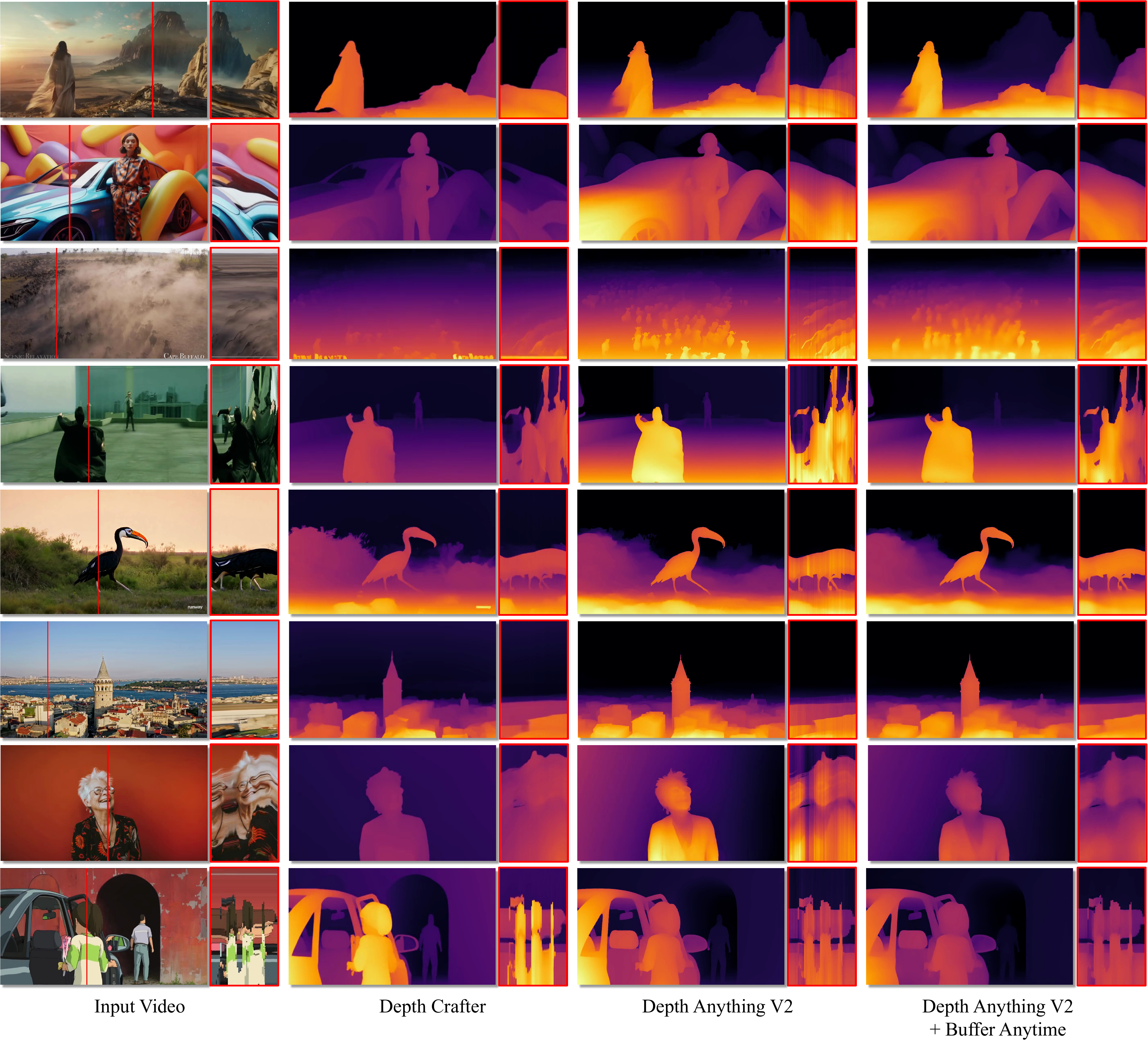}
  \vspace{-20px}
  \caption{\textbf{Qualitative comparison on Video Depth Estimation}. For better visualization, we also show the time slice on the red lines of each video on their right side. Our model keeps the structure details shown in the image model results while achieving smoother performance on the time axis. }
  \label{fig:depth}
  \vspace{-15px}
\end{figure*} 

All of our experiments are conducted on NVIDIA H100 GPUs with 80GB memory. Based on memory constraints, we set the maximum sequence length to 110 frames for depth estimation and 32 frames for normal estimation. We collect approximately 200K videos for training, with each clip containing 128 frames. We use the AdamW~\cite{AdamW} optimizer with learning rates of $10^{-4}$ for depth and $10^{-5}$ for normal estimation. We train on 24 H100 GPUs with a total batch size of 24. The entire training process takes approximately one day for 20,000 iterations.

\subsection{Video Depth Estimation Results}
We evaluate our method on the benchmark provided by DepthCrafter~\cite{DepthCrafter}, which adapts standard image depth metrics for video evaluation. For each test video, the evaluation first solves for a global affine transformation (offset and scale) that best aligns predictions to ground truth across all frames, then computes metrics on the transformed predictions. We report three metrics: Mean Absolute Relative Error (\textit{AbsRel}), the percentage of pixels within 1.25× of ground truth ($\delta_1$), and optical-flow-based smoothness error (\textit{OPW}), defined similarly to our smoothness loss. We evaluate on three datasets: ScanNet~\cite{Scannet} (static indoor scenes), KITTI~\cite{Kitti} (street views with LiDAR depth), and Bonn~\cite{Bonn} (dynamic indoor scenes), using the same test splits as DepthCrafter.

As shown in Tab.~\ref{tab:depth}, our model significantly improves upon its backbone, Depth Anything V2~\cite{DepthAnythingV2}, in both quality and temporal smoothness. Notably, we achieve comparable performance to DepthCrafter, the current state-of-the-art trained on large-scale annotated video datasets, particularly on ScanNet and KITTI datasets. We also demonstrate qualitative comparisons in Fig.~\ref{fig:depth}. Where our model produces more visually stable results than Depth Anything V2, while successfully preserving the structure of the image model prediction.

\begin{figure*}[t]
  \centering
  \includegraphics[width=\linewidth]{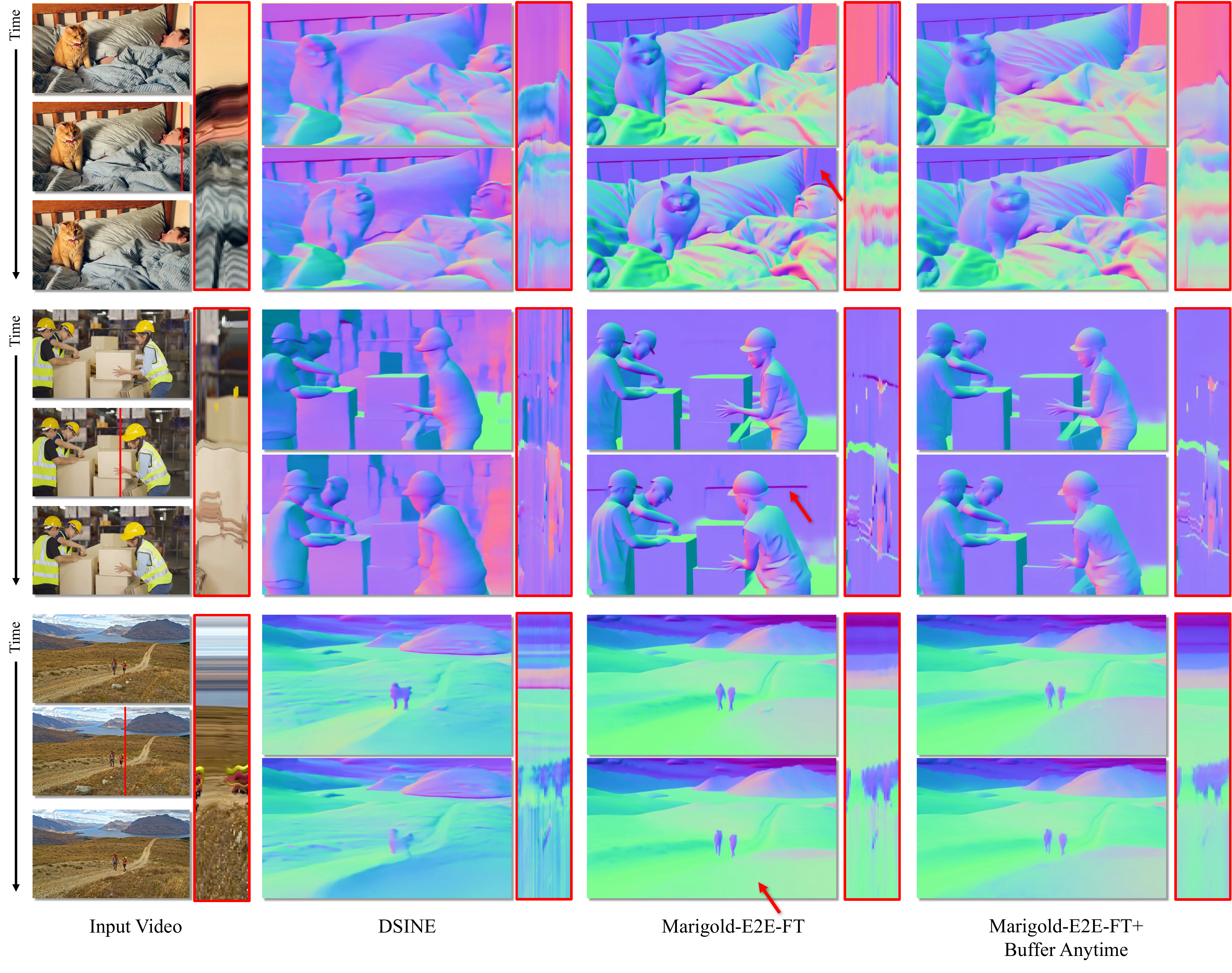}
  \vspace{-20px}
  \caption{\textbf{Qualitative comparison on Video Normal Estimation}. We show the same time slice as in the depth estimation results, and two predicted frames of each model corresponding to the  input frames on the first and third lines. Our model successfully removes the inconsistency from the image model as pointed by the red arrows, achieving smoother results in the time slices. }
  \label{fig:normal}
  \vspace{-13px}
\end{figure*} 

\subsection{Video Normal Estimation Results}
In the absence of a standard video normal estimation benchmark, we establish our evaluation protocol based on the image-level metrics from~\cite{DSINE}. We select two datasets containing continuous frames: Sintel~\cite{Sintel} (synthetic dynamic scenes) and ScanNet~\cite{Scannet}. For each scene, we uniformly sample 32 frames as test sequences. We evaluate using three image-based metrics: mean and median angles between predicted and ground truth normals, percentage of predictions within 11.25° of ground truth, plus the video smoothness metric from our depth evaluation.

Results in Tab.~\ref{tab:normal} show that our model maintains performance comparable to the image backbone on per-frame metrics while significantly improving temporal smoothness. The limited improvement in image-based metrics is expected, as these metrics primarily assess per-frame accuracy rather than temporal consistency. The substantial improvement in the smoothness metric demonstrates our model's ability to generate temporally coherent predictions, as visualized in Fig.~\ref{fig:normal}.

\subsection{Ablation Study}
We conduct ablation studies on KITTI depth estimation to validate our design choices. We compare our full model against four variants: (1) \textit{Ours $\omega_{\mathrm{reg.}}=0.1$} and (2) \textit{Ours $\omega_{\mathrm{reg.}}=3$} use different regularization loss weights; (3) \textit{Ours no mask} omits the optical flow masking; and (4) \textit{Ours all frames} applies regularization to all frames instead of a single random frame. Our full model outperforms the first three variants, validating our design choices. Interestingly, \textit{Ours all frames} shows similar performance to our standard model, suggesting that single-frame regularization sufficiently maintains alignment with the image prior.

\section{Discussion}

In this work, we present a zero-shot framework for video geometric buffer estimation that eliminates the need for paired video-buffer training data. By leveraging state-of-the-art single-view priors combined with optical flow-based temporal consistency, our approach achieves temporally stable and coherent results that match or surpass those of methods trained on large-scale video datasets, as demonstrated in our experiments.

While our approach highlights the power of combining image model priors with optical flow smoothness guidance, there are areas for improvement. First, as our model builds upon image model priors, it may struggle in extreme cases where the backbone model completely fails. Second, while optical flow provides smoothness and temporal consistency between adjacent frames, it only account for correlations across continuous frames. It may fail to, for instance, capture consistent depth information for objects that temporarily leave and re-enter the scene. To tackle these problems, we believe promising future directions are to incorporate large-scale image models with limited video supervision for a hybrid training, or to develop more sophisticated cross-frame consistent guidance (e.g. losses defined in 3D space).

In summary, we propose this framework as a promising step toward reducing reliance on costly video annotations for geometric understanding tasks, offering valuable insights for future research in video inversion problems.

\paragraph{Acknowledgements.} This work was done when Zhengfei Kuang and Tianyuan Zhang were interns at Adobe Research.
\clearpage
{
    \small
    \bibliographystyle{ieeenat_fullname}
    \bibliography{main}
}
\clearpage
\setcounter{page}{1}
\maketitlesupplementary

\section{More Video Results}
In addition to the qualitative comparisons in the paper, we provide more animated results in our supplementary website for better visualization of the prediction quality.

\section{More Implementation Details}
All models are implemented in PyTorch~\cite{Pytorch}. We utilize the official implementations of Depth Anything V2~\cite{DepthAnythingV2} and Marigold-E2E-FT~\cite{E2EFT}, adapting temporal blocks from the UnetMotion architecture in the Diffusers~\cite{Diffusers} library.
Experiments are conducted on NVIDIA H100 GPUs with 80GB memory. Due to memory constraints, we limit the maximum sequence length to 110 frames for depth estimation and 32 frames for normal estimation.

For training, we use a dataset of approximately 200K videos, with each clip containing 128 frames. We employ the AdamW~\cite{AdamW} optimizer with learning rates of $10^{-4}$ and $10^{-5}$ for depth and normal estimation, respectively. Training begins with a 1,000-step warm-up phase, during which the learning rate increases linearly from 0 to its target value. The training process runs on 24 H100 GPUs with a total batch size of 24 and incorporates Exponential Moving Average (EMA) with a decay coefficient of 0.999. The complete training cycle requires approximately one day to complete 15,000 iterations.

\subsection{Details of the Deferred Back-Propagation}


In our normal model, we employ deferred back-propagation as proposed by Zhang et al.~\cite{ARF} to reduce memory consumption. Algorithm~\ref{alg:deferred} outlines the detailed implementation steps. Notably, the gradients obtained by back-propagating $\mathcal{L}_{def}$ are equivalent to those computed from the pixel-wise loss function $\mathcal{L}_{pix}$ across all decoded frames:
\begin{align}
\frac{\partial \mathcal{L}_{def}}{\partial \theta} &= \frac{\partial \frac{1}{K}\sum_{k}  \texttt{Sum}(\texttt{SG}(\bm{g_k}) \cdot \bm{z_k})}{\partial \theta} \\
&= \frac{1}{K}\sum_{k}\bm{g_k} \cdot \frac{\partial \bm{z_k}}{\partial \theta} \\
&= \frac{1}{K}\sum_{k}\frac{\partial \mathcal{L}_{pix}(\mathcal{D}(\bm{z}_k))}{\partial \bm{z}_k} \cdot \frac{\partial \bm{z_k}}{\partial \theta} \\
&= \frac{1}{K}\frac{\partial \sum_{k}\mathcal{L}_{pix}(\mathcal{D}(\bm{z}_k))}{\partial \theta}
\end{align}

\RestyleAlgo{ruled}
\SetKwInput{KwInput}{Input}
\SetKwInput{KwParameter}{Parameter}
\SetKwInput{KwOutput}{Output}
\SetKwComment{Comment}{/* }{ */}

\begin{algorithm}[hbt!]
\caption{Deferred Back-Propagation}\label{alg:deferred}
\KwParameter{Trained model $f_{\theta}$, image decoder $\mathcal{D}$, frame number $K$, chunk size $C$, }
\KwInput{Input frames $\bm{I}_{1,...,K}$, loss function defined on the decoded frames $\mathcal{L}_{pix}$.}
\KwOutput{Deferred back-propagation loss $\mathcal{L}_{def}$}
$\mathcal{L}_{def} \gets 0$\;
$\bm{z}_{1,...,K}\gets f_{\theta}(\bm{I}_{1,...,K})$\;
\For{$ch$ in \texttt{Range}(\texttt{start}=1, \texttt{end}=K, \texttt{step}=C)}
{
    \Comment{Generate chunk prediction} 
    $\bm{z}^{ch} \gets \bm{z}_{ch,...,ch+C-1}$\;
    $\bm{\mathcal{G}}^{ch} \gets \mathcal{D}(\bm{z}^{ch})$\; 
    
    \Comment{Loss on decoded frames} 
    $l \gets \mathcal{L}_{pix}(\bm{\mathcal{G}}^{ch})$\;
    $\bm{g}^{ch} \gets \texttt{Autograd}(l, \bm{z}^{ch})$\;
    \Comment{SG means stop gradient} 
    $\mathcal{L}_{def} \gets \mathcal{L}_{def} + \frac{1}{K}\texttt{Sum}(\texttt{SG}(\bm{g}^{ch}) \cdot \bm{z}^{ch})$ \; 
}
\Return $\mathcal{L}_{def}$
\end{algorithm}


\subsection{Details of the Optical Flow Based Stabilization}
Algorithm~\ref{alg:stable} presents the pseudo-code for our optical flow based stabilization loss calculation. The loss is computed separately for forward optical flow (previous frame to next frame) and backward flow (next frame to previous frame), then combined together. This stabilization algorithm is applied to both depth and normal models. In our experiments, we set the threshold $\tau_c$ to $\frac{\log 2}{2} = 0.34$.


\begin{table}[t]
\centering

{\small \begin{tabular}{l ccc}
\toprule
Method & AbsRel $\downarrow$  & $\delta_1$ $\uparrow$ & OPW $\downarrow$  \\
\midrule
Ours $\mathcal{L}_1$ & 
0.123 & 0.856 & 0.043 \\
Ours w/o fine-tuning & 
0.121 & 0.859 & 0.040 \\
Ours & 
 \rfirst{0.119} & \rfirst{0.865} & \rfirst{0.038} \\
\midrule
Ours with DepthCrafter & 
0.112 & \rfirst{0.884} & \rfirst{0.062} \\
DepthCrafter~\cite{DepthCrafter} &
\rfirst{0.110} & 0.881 & 0.111 \\
\bottomrule
\end{tabular}}

\caption{\textbf{Additional Ablation Study on KITTI depth estimation.} Our model outperforms both variants (\textit{Model with $\mathcal{L}_1$} and \textit{Model w/o fine-tuning}), and when trained on DepthCrafter frames (\textit{Model with DepthCrafter}), achieves comparable performance to DepthCrafter itself.}
\label{tab:ablation_2}

\end{table}

\begin{algorithm*}[tb!]
\caption{Calculating Stabilization Loss}\label{alg:stable}
\KwParameter{Video optical flow model $\mathcal{O}$, frame number $K$, cycle-validation threshold $\tau_c$}
\KwInput{Predicted geometric buffers $\bm{\mathcal{G}}^{pred}_{1,...,K}$, input frames $\bm{I}_{1,...,K}$}
\KwOutput{Stabilization loss $\mathcal{L}_{stable}$}
\texttt{\\}
\Comment{Calculate Optical Flow Maps}
$\bm{O}_{fwd} \gets \mathcal{O}(\texttt{src}=\bm{I}^{pred}_{1,...,K-1}, \texttt{dst}=\bm{I}^{pred}_{2,...,K})$ \Comment*[r]{Shape: $(K-1) \times 2 \times H\times W$}
$\bm{O}_{bwd} \gets \mathcal{O}(\texttt{src}=\bm{I}^{pred}_{2,...,K}, \texttt{dst}=\bm{I}^{pred}_{1,...,K-1})$ \Comment*[r]{Shape: $(K-1) \times 2 \times H\times W$}
\texttt{\\}
\Comment{Calculate Cycle-Validation Masks}
$\bm{\mathcal{M}}^{cyc}_{fwd} \gets \texttt{Where}_{\bm{x}\in\bm{I}_{2,...,K}}(||\bm{O}_{fwd}(\bm{O}_{bwd}(\bm{x})) - \bm{x}||_2 < \tau_{c})$ \Comment*[r]{Shape: $(K-1) \times H\times W$}
$\bm{\mathcal{M}}^{cyc}_{bwd} \gets \texttt{Where}_{\bm{x}\in\bm{I}_{1,...,K-1}}(||\bm{O}_{bwd}(\bm{O}_{fwd}(\bm{x})) - \bm{x}||_2 < \tau_{c})$ \Comment*[r]{Shape: $(K-1) \times H\times W$}
\texttt{\\}
\Comment{Calculate Edge-Based Masks}
$\bm{E} \gets \texttt{CannyEdge}(\bm{\mathcal{G}}^{pred}_{1,...,K})$ \Comment*[r]{Shape: $K \times H\times W$}
$\bm{E} \gets \texttt{Dilate}(\bm{E}, \texttt{kernel\_size}=3)$\;
$\bm{\mathcal{M}}^{edge} \gets \texttt{Where}_{\bm{x}\in\bm{I}_{1,...,K}}(\bm{E}(\bm{x})=0)$ \Comment*[r]{Shape: $K \times H\times W$}
\texttt{\\}
\Comment{Calculate Stabilization Loss}
$\bm{\mathcal{M}}^{fwd} \gets \bm{\mathcal{M}}_{cyc}^{fwd} \wedge \bm{\mathcal{M}}^{edge}_{2,..,K}$\;
$\bm{\mathcal{M}}^{bwd} \gets \bm{\mathcal{M}}_{cyc}^{bwd} \wedge \bm{\mathcal{M}}^{edge}_{1,..,K-1}$\;
$\bm{\mathcal{L}}_{stable}^{fwd} \gets \frac{1}{(K-1)HW} \cdot | (\texttt{Warp}(\bm{\mathcal{G}}^{pred}_{1,...,K-1}, \bm{O}^{fwd}) - \bm{\mathcal{G}}^{pred}_{2,...,K}) \cdot \bm{\mathcal{M}}^{fwd} |_1 $\;
$\bm{\mathcal{L}}_{stable}^{bwd} \gets \frac{1}{(K-1)HW} \cdot | (\texttt{Warp}(\bm{\mathcal{G}}^{pred}_{2,...,K}, \bm{O}^{bwd}) - \bm{\mathcal{G}}^{pred}_{1,...,K-1}) \cdot \bm{\mathcal{M}}^{bwd} |_1$\;
$\mathcal{L}_{stable} \gets \frac{1}{2}(\bm{\mathcal{L}}_{stable}^{fwd} + \bm{\mathcal{L}}_{stable}^{bwd})$\;
\Return $\mathcal{L}_{stable}$.
\end{algorithm*}

\section{Additional Ablation Studies}
We extend our ablation studies beyond the main paper by comparing our model with additional variants:
\textit{Model with $\mathcal{L}_1$} replaces $\mathcal{L}_2$ with $\mathcal{L}_1$ for the affine-invariant relative loss in the depth model; 
\textit{Model w/o fine-tuning} maintains a fixed refinement network from the backbone model while training only the temporal layers.
Additionally, we evaluate an enhanced version utilizing "oracle" knowledge:
\textit{Model with DepthCrafter} incorporates a single frame from DepthCrafter~\cite{DepthCrafter} prediction per iteration as regularization guidance.

As shown in Table~\ref{tab:ablation_2}, our model demonstrates superior performance compared to the first two variants, validating the effectiveness of both our architectural and loss function designs. The \textit{Model with DepthCrafter} achieves better results that comparable to DepthCrafter itself, suggesting potential for future improvements through enhanced image priors.


\end{document}